# Capabilities: An Ontology

John Beverley[1,2,3,*], David Limbaugh[1], Eric Merrell[1], Peter M. Koch[1,3], Barry Smith[1,2,3]

[1] National Center for Ontological Research, Buffalo, NY, USA
[2] University at Buffalo, Department of Philosophy, Buffalo, NY, USA
[3] Institute for Artificial Intelligence and Data Science, University at Buffalo, NY, USA
[4] Villanova University, Department of Philosophy, Pennsylvania, USA

**Abstract**

In our daily lives, as in science and in all other domains, we encounter huge numbers of dispositions (tendencies, potentials, powers) which are realized in processes such as sneezing, sweating, shedding, melting. Among this plethora of what we can think of as 'mere dispositions' is a subset of dispositions in whose realizations we have an interest – a car responding well when driven on ice, a rabbit's lungs responding well when it is chased by a wolf, and so on. We call the latter 'capabilities' and we attempt to provide an ontological account of what capabilities are that is of sufficient generality to serve a variety of purposes, for example by providing a useful extension to ontology-based research in areas where capabilities data are being collected in siloed fashion.

## 1. Introduction

Capabilities are everywhere. There are software capabilities, gardening capabilities, intellectual capabilities; there are innate capabilities and capabilities gained through practice; capabilities of individuals, capabilities of organizations, and capabilities of machines. There is a large literature on capabilities of different types. But there is not, to our knowledge, any serious attempt to create an account of capabilities that is of sufficient generality for use as a starting point for the formulation of mutually consistent definitions of terms referring to the various subspecies of capability that have been identified in different areas. An account of this sort would then be useful especially in those areas where capability identification can be an urgent need – in emergency management, for example, where responsible organizations need to have rapid access to relevant experts who can advise and assist following a disaster [1]. Sample areas in which capability identification are of importance include:

---

*Proceedings of the Joint Ontology Workshops (JOWO) - Episode X: The Tukker Zomer of Ontology, and satellite events co-located with the 14th International Conference on Formal Ontology in Information Systems (FOIS 2024), July 15-19, 2024, Enschede, The Netherlands.*

* Corresponding author.
johnbeve@buffalo.edu (J. Beverley); davidglimbaugh@gmail.com (D. Limbaugh); ontomerrell@gmail.com (E. Merrell); petermkoch@gmail.com (P. M. Koch); ifomis@gmail.com (B. Smith)

*Medicine.* When a patient is admitted to a hospital, medical professionals with salient capabilities must be identified and made available. There are relevant capabilities also on the side of the patient, and medical professionals may need to communicate information for example regarding a patient's interests and values, as well as their responses to medication and treatment [2]. A patient's responsiveness to her doctor's instructions, for example, is a capability of the patient that is associated with the satisfaction of her interests, including her interest in survival [3]. Epidemiologists and pathologists must often communicate information concerning the capabilities of medications. Remdesivir, for example, was developed to treat hepatitis C, repurposed to treat Ebola, and may be effective at treating COVID-19 [4, 5]. Such creative exploration of potential new uses of existing drugs, which is at the heart of rational drug design, is exploration in the realm of capabilities.

*Industrial manufacturing.* We find in the industrial domain myriad examples of ways in which capabilities-related data play a vital role. Materials – both kinds of materials and individual batches – are selected as inputs into manufacturing processes based on capability-related properties such as availability, malleability, elasticity, electrical resistance, and so forth. Maintenance data are used to document the effects of different maintenance regimes on the ways the capabilities of equipment and products evolve over time. Supply chain data include information on the capabilities (reliability, flexibility, financial stability, precision of work) of potential suppliers [6, 7].

*Defense and security.* National security decisions require that we have an accurate understanding of the capabilities[2] not only of ourselves and of our allies but also of our actual and potential adversaries [8], and of the complex capabilities the need for which arises in specific mission contexts. Mission-relevant capabilities may be borne not only by single warfighters but also by collectives thereof, and new capabilities may arise as a result of how collections are configured [9]. One infantryman, for example, may be trained to leverage GPS targeting equipment, another to operate surface-to-air missiles. Together they have the capability to engage enemy aircraft. An air force unit trained in combination with a unit of ground troops has the capability to engage in joint air-and-land assaults. This idea was developed into a form of *capabilities engineering* in the framework of what the Chairman of the Joint Chiefs of Staff defined as 'Joint Capability Areas', which are "DoD capabilities functionally grouped to support capability analysis, strategy development, investment decision making, capability portfolio management, and capabilities-based force development and operational planning." [10, 11]

*Education and training*. Capabilities play a central role in the domain of education, which is a complex of activities designed to produce persons with cognitive capabilities at successively higher levels of sophistication, and to produce at the same time persons with the capabilities required by educators themselves at successively higher tiers in the hierarchy of educational institutions. An organization's training activities – whether devoted to the training of Olympic athletes or of Customs and Border Protection field agents – have a similar goal. Organizations have an interest in hiring talented individuals to further organizational goals. Individuals have an interest in acquiring the sorts of capabilities needed to address an organization's needs.

---

[2] And also vulnerabilities – a topic we will address in a sequel to this communication.

Testing both students and trainees serves the purpose of measuring the degree to which they can realize the capabilities they are being schooled or trained to acquire. Acquisition of such capabilities is then signaled through diplomas and certifications. What holds for trainees holds equally for artifacts, where we also find batteries of tests and other mechanisms designed to validate improvements or identify deficits in artifact capabilities.

## 2. Characterizations of Capabilities

Just as there is a wide range of contexts in which the term 'capability' is used, so there is a wide range of definitions of this term. In the literature on education, for example, 'capability' has been defined as what an individual "could achieve … an endowment … realized by the acquisition of skills" or as what an individual "can (or has learned to) do." [12]. In engineering, a capability has been defined as an "activity that a resource could undertake and the related outputs" [13]. In economics, it has been defined as "the set of alternatives … from which an individual can actually choose … given certain resources and skills" [14]. For the DOD, a capability is "the ability to achieve a desired effect under specified standards and conditions through combinations of means and ways to perform a set of tasks" [15, 16, 17].

One important set of difficulties standing in the way of consistent and coordinated treatment of capabilities data turns on the abundance of (near) synonyms of the term 'capability' with which relevant data are labelled.[3] For example, given that the term 'ability' is often used as a near synonym of 'capability' the just-mentioned DOD definition is potentially circular. The most influential philosophical writings on this topic characterize capabilities using phrases such as "What [people are] actually able to do and to be?" [18] Use of phrases such as 'is able to do' means that these definitions, too, border on circularity.

There are also differences in range of application of the term 'capability'. Philosophical accounts, for example, tend to restrict capabilities to what *persons* or *organisms* can do, despite the plausibility of the idea that organizations, buildings, and machines can bear capabilities. One of us has provided the following more inclusive philosophical characterization of capability as "a disposition which can be exercised more or less well and which is good for the bearer or the bearer's wider community when realized" [2]. This definition makes explicit what is implicit across the preceding definitions: capabilities aim to satisfy some goal or interest and that such satisfaction may manifest in degrees.

### 2.1. Hallmarks of Capabilities

The above characterizations suggest the following *hallmarks of capabilities*, namely that:

1. they are found across almost all domains,
2. they are closely related to what an entity can do,
3. they may or may not be realized,

[3] Examples include: ability, capacity, competence, expertise, facility, know-how, proficiency, skill.

4. they are borne by an entity or group of entities,
5. they may be realized to different degrees,
6. they are associated with the satisfaction of some goal or interest.

We believe that an accurate ontological characterization of capabilities should reflect these hallmarks. In what follows, we will propose such a characterization, taking inspiration from the inclusive philosophical characterization of capabilities in terms of what is good for a bearer or its community [2], but adding precision, and pointing to applications which reveal the need to broaden the account beyond the focus of what is good for the bearer.

## 3. Basic Formal Ontology

In formulating our proposed definition, we adopt Basic Formal Ontology (BFO) as our starting point, drawing on the fact that BFO already contains formal and natural language definitions of many of the terms and relations we will need in our analysis. BFO is a top-level ontology, which is to say, it is a taxonomy of types of entities of a highly general sort – represented by terms such as **object**,[4] **quality**, **process** – which are used in practically all domains of human activity. It contains both natural language and logical definitions of such general terms, as well as definitions of relations – such as *participates in*, *is located at*, *is part of* – by which entities referred to by using these terms may be linked together in reality.

BFO was developed originally to enable the consistent expression of data across multiple disciplines of biology and medicine. Nowadays, major users of BFO include the developers of the ontologies in the Open Biological and Biomedical Ontologies (OBO) Foundry [19], in the Industrial Ontologies Foundry (IOF) [20], and in the Common Core Ontologies (CCO) ecosystem of military- and security-related ontologies [21]. BFO is used in some 600 open-source ontology initiatives [22], where it provides a common, domain-neutral framework for ontologies developed not only by scientists working in different domains but also by government, financial and industrial organizations. BFO has been accepted by the International Standards Organization as ISO/IEC 21838 standard top-level ontology [23], and as of January 2024 BFO and CCO have been adopted as baseline standards for all formal ontology development in the DOD and Intelligence Community [24].

### 3.1. The Basic Formal Ontology Hierarchy

Terms in BFO and in BFO-conformant ontologies represent classes of instances that share important features. The highest division in BFO's class taxonomy is that between **occurrent** and **continuant**. **Occurrents** are either (i) extended in time in such a way as to have temporal parts (phases, stages, ...) which are extended over time; these are either **processes** or the **temporal regions** in which **processes** occur, or (ii) they are the **boundaries**, for example beginnings and endings, of such **processes** or of the associated **temporal intervals**, or (iii) they are **spatiotemporal regions** linking **temporal regions** to space on the **continuant** side of BFO.

---

[4] We use **bold** for terms referring to ontology classes; for example using '**capability**' (or sometimes '**capabilities**') to represent the class of capabilities; we use *italics* for relational terms, such as *continuant part of.*

**Continuants**, on the other hand, (i) lack temporal parts, (ii) endure self-identically through time, and (iii) are typically able to change their qualities and to gain and lose parts in the course of their existence. **Continuants** and **occurrents** are tied together by the fact that the former *participate in* **processes** as when your mother *participates in* an instance of jaywalking. The **continuant** class has three subclasses, as illustrated in **Figure 1**. First, an **independent continuant** is a **continuant** that does not depend on anything for its existence [25].[5] A baseball is an **independent continuant.** The mass and shape of the baseball are **specifically dependent continuants**. Each instance of the latter depends for its existence on some specific instance of the former.

**Independent continuant** has two BFO sub-classes, namely **material entity** and **immaterial entity**, which consist, respectively, of entities having and entities lacking matter as parts.[6] Subclasses of **material entity** include **objects**, such as Mick Jagger, and **object aggregates**, such as the Rolling Stones.[7] Subclasses of **immaterial entity** include **spatial regions** and **sites** (for example: the Rafah Crossing, the Mariana Trench), as well as various continuant boundary entities (for example, the Equator).

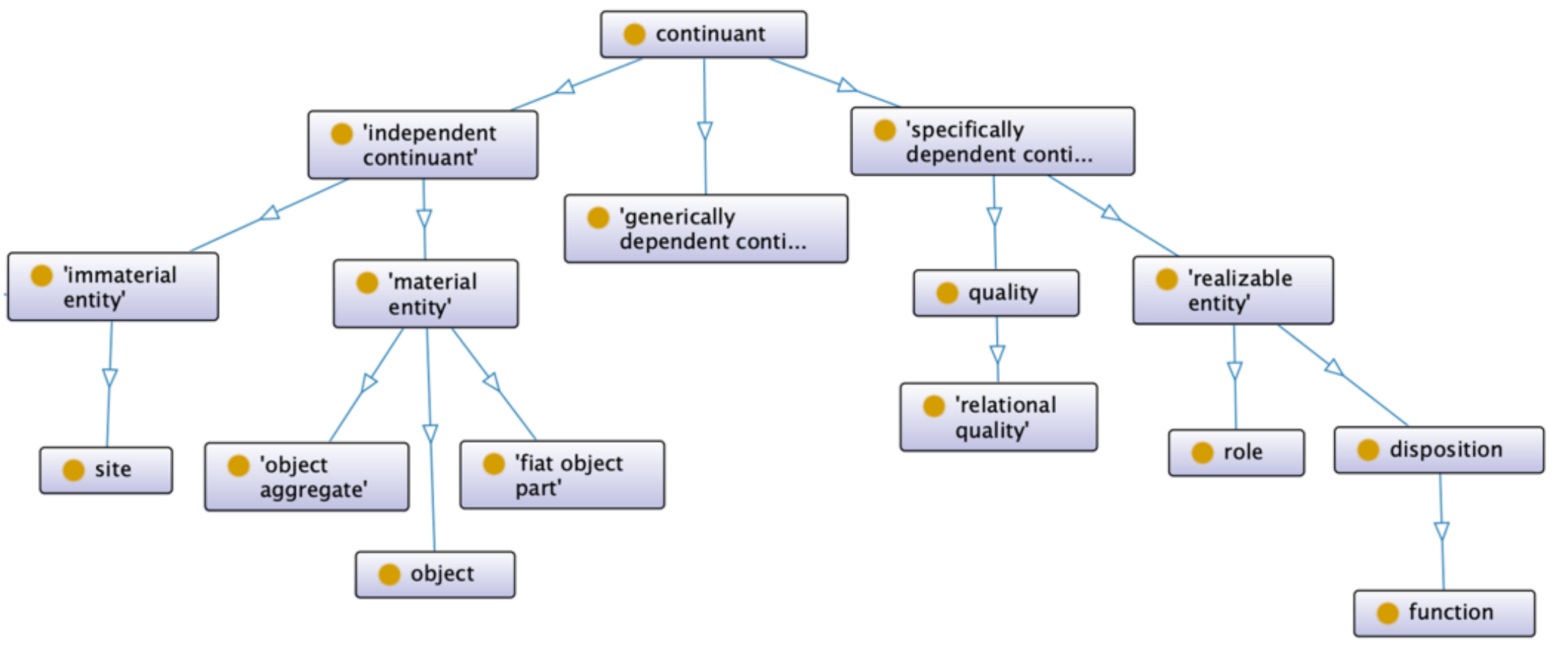

***Figure 1****: Portion of the Basic Formal Ontology Taxonomy*

Certain instances of **specifically dependent continuant** are fully manifested whenever they manifest at all, such as color, shape, or mass; these are instances of the class **quality**. **Realizable entities**, in contrast are those **specifically dependent continuants** which are marked by the fact that they need not manifest when they exist. If an apple can float in water, it possesses this ability even when it is not in water. **Realizable entities**, in sum, must be realiz*able*, but they are not in every case realized. Two major subclasses of **realizable entity** recognized by BFO are **dispositions** and **roles**. **Dispositions** are **realizable entities** that are internally grounded. This means that, for a **disposition** to begin or cease to exist, its bearer must undergo a physical change.[8] For example, a portion of salt may lose its solubility, but only if it undergoes some physical change to its lattice structure [29]. **Roles**, in contrast, must be externally grounded. This means that the gain or loss of a **role** – for example your promotion to a new rank – does not imply change in your physical makeup.

[5] *Depends on* here is to be understood in the ontological sense. Thus it holds between two relata *x* and *y* when the former is such that it cannot exist unless the latter exists [26].
[6] See [27] for a discussion of BFO's mereological commitments.
[7] We leave aside here the BFO treatment of fiat object parts, such as: your head, or the Southern Hemisphere.
[8] See 'intrinsic' and 'extrinsic' in [28].

The class **disposition** has a single subclass recognized in BFO, namely **function,** which is (roughly) a **disposition** that reflects the reason for the existence of its bearer.[9] **Functions** fall into two main types. On the one hand are functions of artifacts, which are entities designed and intentionally created to realize some function. On the other hand are **functions** of biological entities. A human heart bears a **function** to pump blood. The heart evolved in virtue of the fact that an organ fulfilling such a purpose was instrumental in keeping our ancestors alive.

A **generically dependent continuant** is (roughly) a copyable pattern. A pattern of this sort exists only if it is concretized in some continuant bearer; but it is not dependent on any specific bearer, because it may be copied (for example through being transmitted) from one bearer to another. Here, again, there are two major subclasses. On the one hand are information artifacts, such as digital blueprints or embroidery patterns. On the other hand are biological examples, such as nucleotide sequences. **Table 1** provides the definitions or (in the case of primitives) elucidations of terms in the BFO hierarchy.[10]

**Table 1**
Definitions/elucidations of selected BFO classes

| BFO Class | Elucidation/Definition |
|---|---|
| ***Continuant*** | An entity that persists, endures, or continues to exist through time while maintaining its identity. |
| ***Independent Continuant*** | A continuant which is such that there is no *x* such that it specifically depends on *x* and no *y* such that it generically depends on *y*. |
| ***Specifically Dependent Continuant*** | A continuant which is such that (i) there is some independent continuant *x* that is not a spatial region, and which (ii) specifically depends on *x*. |
| ***Generically Dependent Continuant*** | An entity that exists in virtue of the fact that there is at least one of what may be multiple copies. |
| ***Material Entity*** | An independent continuant that at all times at which it exists has some portion of matter as continuant part. |
| ***Object*** | A material entity which manifests causal unity and is of a type instances of which are maximal relative to the sort of causal unity manifested. |
| ***Object Aggregate*** | A material entity consisting exactly of a plurality (≥1) of objects as member parts which together form a unit. |
| ***Quality*** | A specifically dependent continuant that, in contrast to roles and dispositions, does not require any further process in order to be realized. |
| ***Realizable Entity*** | A specifically dependent continuant that inheres in some independent continuant which is not a spatial region and is of a type some instances of which are realized in processes of a correlated type. |
| ***Role*** | A realizable entity that exists because there is some single bearer that is in some special physical, social, or institutional set of circumstances in which this bearer does not have to be, and is not such that, if it ceases to exist, then the physical make-up of the bearer is thereby changed. |

[9] BFO adopts an account of **function** broadly in conformity with Millikan's 'historical' account of what she calls 'proper functions' [30]. For further details see [31] and, specifically on biological functions, [32].

[10] Elucidations are descriptions provided to help fix the referent of primitive terms. Definitions express individually necessary and jointly sufficient conditions for an entity to be an instance of the class defined.

| | |
|---|---|
| ***Disposition*** | A realizable entity such that if it ceases to exist, then its bearer is physically changed, and its realization occurs when and because this bearer is in some special physical circumstances, and this realization occurs in virtue of the bearer's physical make-up. |
| ***Function*** | A disposition that exists in virtue of the bearer's physical make-up and this physical make-up is something the bearer possesses because it came into being, either through evolution (in the case of natural biological entities) or through intentional design (in the case of artefacts), in order to realize processes of a certain sort. |
| ***Occurrent*** | An entity that unfolds itself in time or is the start or end of such an entity or is a temporal or spatiotemporal region. |
| ***Process*** | An occurrent that has some temporal proper part and for some time has a material entity as participant. |

## 4. Defining 'Capability'

An SUV is capable of ferrying passengers and luggage. Were the SUV to cease to exist, so too would its capabilities. A dependence of this sort on some **material entity** exists for all capabilities. This tells us that **capabilities** in BFO are best described as **specifically dependent continuants**. **Capabilities** are, moreover, instances of what BFO calls **realizable entities**. If they are realized, then it is always through some **process**. This fact provides the link between **capabilities**, on the one hand, and **processes** on the other.

The choice of whether **capability** best falls under **role** or **disposition** turns on whether **capabilities** are internally or externally grounded – or in other words on whether they are grounded in their bearers or in something – for example some authority – in their wider environment. A litmus test for determining whether a **realizable entity** is a **role** or a **disposition** is: does the gain or loss of the **realizable entity** require some physical change to its bearer? [25] If yes, then it is internally grounded and so a **disposition**. If no, then it is externally grounded and so a **role**. A portion of steel may gain or lose its ability to cut through softer metals, either of which would result in physical change in the steel itself. An untenured assistant professor may gain tenure when a certain document is signed without herself being physically changed in any way.

The examples of capabilities that we have presented above fall always on the **disposition** side. A student who learns to code in C++ has acquired a **capability** to do so, the loss of which would involve some material change localized in her brain. That said, there seem to be plausible uses of the term 'capability' on the **role** side also. As an example, let us suppose that you are promoted on a temporary basis because your superior has called in sick. People might then say that you now have the capability to issue orders, where '*x* has the capability to do *y*' means something like '*x* is allowed' or 'has the authority to do *y*'. While this certainly involves a sense of 'capability' that is of importance to the full understanding of its natural language use, this meaning is orthogonal to the sense at issue here.

**Dispositions** and **roles** are nonetheless in many cases intimately related. This is because, even though they are ontologically distinct, they are often instantiated together in a single bearer. If you are appointed to the **role** of floor sweeper, then you will very soon, as you proceed to realize your new **role**, acquire **capabilities** associated with floor sweeping. As another example, suppose that I have set aside a certain flathead screwdriver for use in opening paint cans. Then it seems plausible to understand that screwdriver as having both a **role** and a **disposition** [31, 33]. And this is as it should be. For the **role** borne by the screwdriver is a **role** whose realizations depend in part on **capabilities** borne by the screwdriver. It is in fact *because* of such **capabilities** that the screwdriver can even be considered for the **role** of paint can opener.

One consequence of our account is that, while **immaterial entities** such as **sites** may bear **roles**, because they lack material parts they cannot bear **capabilities**. This is a feature rather than a bug. It makes sense to say that the **site** that is the air space above Utah bears a **role** in, say, U.S. military defense protocol. It makes little sense to say that this entity bears a **capability**. Similarly, it is the surrounding fences, gates, and guards bearing arms which together bear the **capability** to allow and to prevent passage through the Rafah Crossing.[11]

Our proposal also implies that **generically dependent continuants** such as software tools, which also lack material parts, cannot bear **capabilities**. That said, 'software tool' is ambiguous as between portions of code, algorithms, scripts, and so on, on the one hand – which are **generically dependent continuants** – and, on the other hand, the hardware implementing such entities: for example a central processing unit in which a given piece of software is installed. The latter is a **material entity** in BFO terms, and any software tool in the former sense must, if it is to bear **capabilities**, realize those **capabilities** via execution in hardware. Talk of algorithms 'bearing capabilities' is thus best understood as shorthand for statements about capabilities of the hardware-software amalgam which results when the algorithms are being implemented.

### 4.1. Interest In

**Capabilities** are in every case **dispositions** whose realizations are associated with some goal or, more generally with some interest, which an organism or group of organisms has. If you have an *interest in* a thing in the sense at issue here, this implies that whatever affects the thing directly can have a direct or indirect effect on you. To have an interest in something is to *not be a disinterested party*. You have an interest in keeping your neighborhood clean. You do not have an interest in keeping clean the surfaces of passing asteroids. In the relevant sense here, therefore, the meaning of 'interest in' is skew to what we might call the curiosity sense of 'interest' (with negative form *uninterested in*) that is manifested, for example, in scientific research.

Having an interest in the realization of a **capability** is a relation which holds between organisms or groups of organisms and **processes** that are realizations of **dispositions** in some **material entity**. The reference to groups turns on the essentially collective nature of many capabilities. One person alone does not have the capability to dance the foxtrot, for example. It has been argued that language, too, is an essentially collective capability [34].

There is an important special class of interests which arise through the presence of goals. Goals are phenomena of a type characteristic of human organisms in that they go beyond mere survival and reproduction. When someone has a goal, then they have at least a minimal plan to achieve that goal, and at least some **disposition**, however weak or incipient, to realize the plan. If, for example, your goal is to own a Ferrari, then you must have at least a minimal plan to raise the needed funds, and you will then, given your goal, have an interest in **processes** of the type prescribed by that plan. In many such cases, realizing the plan will involve the creation of new **capabilities**. Complex plans, such as launching a space station into orbit, will involve the specification of huge numbers of **processes** chained together at many levels. Many of these **processes** will require the creation of specific **capabilities** on the side of both humans and machines. Many of the latter will be **functions** of the relevant artifacts, namely in those cases where artifacts are brought into being because of a plan to create an entity that can realize processes of a certain sort.

[11] One family of examples of **capabilities** – bodyguard capabilities – are *per accidens* never, or only rarely, realized. The **capability** of your immune system to respond, for example, to a Helmuth infection, may never be realized. The **capability** of parts of your bank's vault designed to lock the vault during a break-in, may never be realized.

Plans and goals aside, there are endlessly many cases where non-human organisms have an interest, though we cannot properly speak of their 'having goals'. There is some residual sense in talking of survival and reproduction as goals of, for example, a nematode worm. But then the features that we associate with having *goals* on the part of humans[12] are absent. Our proposal thus covers cases where interest is and is not related to goals.

We take 'has an interest in' as a primitive relational expression, which means that we cannot provide a non-circular definition [35]. We can however assert necessary conditions for the relation to obtain:

*o has an interest in p* only if

1. *o* is an organism or group of organisms.
2. For some material entity *m* with dispositions $b_i$ and realizations $p_i$: *p* is among these realizations and *p* may causally influence or be influenced by *o*. Examples of *m* are: you, your heart, your home, your loved ones, your enemies. Examples of *p* are: your heart beating, your daughter winning a race.
3. *p* contributes to the survival and reproduction of *o* or to the realization of *o*'s goals.

These conditions allow that one need not be aware that one has interest in realizations of some **process** in order to have a corresponding interest, as is the case when a nematode has an interest in survival, or your grandfather has an interest acetylcholine uptake in his synapses.[13]

### 4.2. Not All Capabilities are Functions

There are similarities between **capabilities** and **functions**. First, any instance of **function** exists in virtue of its bearer's physical makeup, which is a feature shared with **capabilities**. Each instance of **function** is further such that its realizations can be evaluated according to some standard [31]. The realizations of **capabilities**, too, can be evaluated in this way. Your car's capability to transmit in stereo may be realized with a signal of greater or lesser clarity.

But there are also differences. Human beings (in our view) do not have **functions**, but they do have **capabilities,** because they have parts with functions whose realization contributes to their survival. My digestive system, for instance has the **function** to digest food, and therefore I have the **capability** to digest food. Similarly, our flathead screwdriver bears a **function** to drive screws using torque, but it may also, as we saw, bear the **disposition** to open paint cans. And if people have an interest in using a screwdriver in this way, then its **disposition** to open paint cans acquires the status of a **capability**.

[12] For a formalization of these features, compare the treatment of 'plan' in [36].
[13] We discuss this topic in more detail in **4.6**.

Additionally, each instance of **function** is such that we can advert to this **function** in explaining how its bearer came into existence. The presence of a **capability**, in contrast, need have no etiological consequences of this sort.

We can now offer the following definition of **capability**:

> *x* is a **capability** just in case *x* is a **disposition** in whose realization some organism or group of organisms has or had[14] an interest.

### 4.3. All Functions are Capabilities

The above observations motivate our conclusion that not all **capabilities** are **functions**. Nevertheless, all **functions** are **capabilities**. A **function** in BFO is borne either:

1. by a biological entity (a part of some organism), whose **function** is such that the bearer has an *interest in* its realization because the bearer has an interest in its own survival and/or reproduction; or
2. by an artifact that is the product of intentional design by one or more organisms who have an *interest in* the realization of the **function** of the artifact.

In either case there is some organism or group of organisms which has an *interest in* the realization of a **function**, and since the **function** is a **disposition**, they have an *interest* also in the realization of a **disposition**. If *x* is a **disposition** in whose realizations an organism has an *interest*, then *x* is a **capability**. It follows that all **functions** are **capabilities**.

### 4.4. Functions vs Role

Our account helps to disambiguate certain natural language uses of 'function' and 'role'. The term 'functional role', for example, is often used to describe some key individual within an organization who must perform certain activities to ensure that the organization operates properly. In BFO, such activities would involve individuals bearing **roles** of the mentioned sort. But the bearer of such a **role**, for example the mayor of a city, must in addition have the **capabilities** needed to perform the required operations, such as speaking the relevant language. In short, the mayor's 'functional role' is indeed simply a **role**, but one whose realization depends on various **capabilities**, whether pre-existing or acquired on the job.

Consider, similarly, that part of the literature on functions which distinguishes between 'primary' and 'secondary' **functions** [37]. The primary function of a car engine is to enable motion of the car. This it achieves through the realization of secondary functions borne, for example, by piston parts, functions which they realize by combusting gasoline. Here both primary and secondary functions satisfy our definition of **function** *simpliciter*. A secondary function is a **function** whose realization enables an object to perform its primary function. The significance of this idea turns on the fact that one might design engines that accomplish the same primary function by realizing different secondary functions, for example by means of an engine whose parts are designed to combust diesel fuel or methane.

In this way, distinguishing between primary and secondary functions can become part of a strategy to achieve cost-effective artifact design. For example, a car with an internal combustion engine emits considerable heat following combustion. The internal combustion engine bears neither a primary nor a secondary function to produce excess heat. However, because the heat produced is sufficient to warm a car cabin at no additional energy cost, the combustion engine bears a **capability** to produce excess heat. It is recognition of such a **capability** during design of a car heating system that is leveraged to reduce costs by having excess heat diverted into the car cabin. In other words, the primary function of a car heating system is partially enabled by the **capability** of the engine to produce excess heat.

---

[14] On this 'or had' clause see **4.5** below.

### 4.5. Capability Death

When does a **capability** cease to exist? Here we defend an account that is modelled on the account we provide of the end-of-life for **functions**. According to BFO, a **function** exists if and to the extent that there is a real possibility for it to be realized [30]. Whether the relevant possibility exists turns on the intrinsic physical structure of the bearer of the **function**. If the old heater in my attic is undamaged then it continues to have the function of a heater, and this would be so, providing its mechanism survives intact, even if the heater were to be encased in concrete. If, on the other hand, the heater is broken beyond repair, then it thereby loses its **function**. **Functions** may on this view become obsolete, but they still survive as **functions** providing that the bearer has suffered no physical changes of a sort that would preclude realization.

A similar argument, now, can be applied to **capabilities**, resulting in the following constraint:

> If *x* has an *interest in* the realization of **disposition** *y* at time *t*, then *y* is a **capability** at *t* and at all subsequent times during which *y* exists.

A Betamax player, still today, bears a **function** to play Betamax tapes, and this is so even if no one is left who has an *interest in* realizing this **function**. Since every **function** is a **capability,** it follows trivially that the Betamax player bears a **capability** to play Betamax tapes independently of any surviving interest. As a perhaps more extreme example, consider the discovery of an ancient artifact – a stick, of a type used in certain now forgotten rituals of human sacrifice. No one today has an interest in realizations of the specific **dispositions** borne by this specific stick. But the sacrificial high priest had such an interest. Since, in the intervening centuries, the stick is (by assumption) not physically changed, it follows that it still has the corresponding **capabilities**. This is a desirable consequence: among the reasons archeologists have sought intact exemplars of a stick of this type was precisely because of the desire to test their conjectures concerning their **capabilities**.

### 4.6. Capability Birth

**Capabilities** begin to exist either when a **disposition** *begins to exist* in response to the interests of some organism, or when an already existing **disposition** *becomes* a **capability** because an organism acquires an *interest in* its realization. That many **capabilities** are being created *ab ovo* is revealed by annual patent statistics. Pre-existing **capabilities** may be discovered independently, for example by different groups of researchers. Much as the physical basis of a **disposition** underwrites its existence independent of any awareness, so interest in the realizations of a **disposition** underwrites its existence as a **capability** independent of any awareness of this interest. There are many **dispositions**, for example dispositions of many of the cells in our bodies, of which we are not aware. And thus, given that we have an interest in the functioning of these cells, there are many **capabilities** of which we are not aware. Sam's pancreas bears a **function** to produce insulin. Sam has an *interest in* this **function** being realized. But Sam need not have any accompanying knowledge associated with producing insulin.

That **dispositions** are **capabilities** may be discovered. In 1971 an organization called 'Federal Express' was created, with the **function** to deliver parcels. In the year 2000 there was founded a new organization in the Fedex family, organization that has since 2009 been titled 'Fedex Logistics'. FedEx, it transpired, had been providing logistics services for many years, but did not at first realize that it was in possession of a separable logistics **capability** from the exercise of which it could make money [38].

### 4.7. Objections and Responses

One criticism of our proposal is that it would seem to depart from the scientific orientation of BFO because it is not morally neutral. It seems, in some way, to make 'being a capability' rest on a distinction among **dispositions** between the good and the bad. To see that this is not so, consider the case of a human being who is a functional sociopath. He lacks a conscience and is willing and able to kill with impunity. This lack of a conscience is not, from your and our perspective, a **capability**. But from the perspective of a drug cartel mercenary army the underlying **dispositions** are most surely **capabilities**. Whether a **disposition** is or is not a **capability** may in this way depend on the interests of the parties involved.

We note that a parallel concern arises also in the case of **roles**. Many **roles** are by their nature social constructions; they depend on institutions and groups, and new roles – the role of *picciotti d'onore* of the 'Ndrangheta, for example – may be created and assigned in morally neutral or morally loaded ways.[15] Both **role** and **capability** capture complex phenomena limited only by the creativity of human beings. These facts are not in conflict with the scientific basis of BFO. Even so, one may still object that, according to our proposal, **capabilities** may be created *too easily*, almost effortlessly, simply by someone's having an *interest in* realizations of a certain type. In response, note that there are, importantly, constraints on the sorts of **dispositions** in which we can have an interest. First, having an interest implies something like having a stake in some object's future destiny – your daughter, for example, but not the twin Earth counterpart of your daughter. Second, *interest in* is always *de re*, and not merely *de dicto*. If I have an *interest in* eliminating a squeak in a door hinge and some substance exists in the world with lubricating **dispositions**, it does not follow that those **dispositions** are now **capabilities**. My interest is in **processes** realizing the elimination of the squeak in this door hinge that are within the realm of what I can bring about causally. Thus, while *interest in* is the ultimate arbiter of whether a **disposition** is a **capability**, the specificity of interest is a delimiting factor.

One may object that it is unreasonable to accept that any disposition, without undergoing any change, can suddenly transform into a **capability**. A change does occur, however; for as a result of this transformation the **disposition** gains a relation to the salient organism or group of organisms that it did not have before. It should, moreover, be no surprise that the class of **capabilities** represents an easily expandable terrain. Our interests are vast and varied. If my enemies have an *interest in* my death and I have a cancerous tumor in my brain, then the tumor bears **dispositions** that my enemies have an *interest in* seeing realized. Under these circumstances my tumor bears **capabilities**. This is as it should be. On the other hand, I have numerous **dispositions** in whose realization no one has an interest, my enemies included, and so these are not **capabilities**. This, too, is as it should be.

---

[15]Note that there are cases of 'apparent' role assignment, as described in this story, told of the philosopher Wittgenstein: "On one walk he 'gave' to me each tree that we passed, with the reservation that I was not to cut it down or do anything to it, or prevent the previous owners from doing anything to it … it was henceforth *mine*." [39]

## 5. Conclusion

We propose an ontological characterization of 'capability' as a first step towards ameliorating some of the problems caused by the uncontrolled use of this term and its manifold synonyms in a variety of domains. We provide a definition for **capability** and elucidation of the *interest in* relation within the BFO context. We argued that the proper placement of **capability** is between the BFO classes **disposition** and **function**. **Functions** are **capabilities**; **capabilities** are **dispositions**. Despite these results, the complexity of **capabilities** warrants further analysis. Though we are not suggesting **capability** be included within the BFO hierarchy, we offer the present characterization of **capabilities** to the BFO community as a resource for modeling this pervasive phenomenon.